%% file: neurips_2026.tex
\documentclass{article}

\usepackage[preprint]{neurips_2026}

\usepackage[utf8]{inputenc} 
\usepackage[T1]{fontenc}    
\usepackage{hyperref}       
\usepackage{url}            
\usepackage{booktabs}       
\usepackage{longtable}      
\usepackage{amsfonts}       
\usepackage{nicefrac}       
\usepackage{microtype}      
\usepackage{xcolor}         
\usepackage{amsmath,amssymb}
\usepackage{enumitem}
\usepackage{graphicx}       

\title{\texttt{vidax}: A Unified JAX Framework for Video Generative Models on Accelerator Meshes}

\author{%
  Congyue Deng \\
  Massachusetts Institute of Technology\\
  \texttt{congyued@mit.edu} \\
}

\begin{document}

\maketitle

\input{sections/0_abstract}
\input{sections/1_introduction}
\input{sections/2_system}
\input{sections/3_models}
\input{sections/4_benchmark}
\input{sections/5_conclusions}

\begin{ack}
This project is supported by the Google TPU Research Cloud (TRC) Program. Congyue Deng also acknowledges the Tayebati Postdoctoral Fellowship.
\end{ack}

{
\small
\bibliographystyle{plain}
\bibliography{references}
}

\appendix
\input{sections/X_appendix}

\end{document}

%% file: sections/0_abstract.tex
\begin{abstract}
Open-source video generative models ship almost exclusively as PyTorch/CUDA reference implementations. This leaves Cloud TPU pods without a production-ready inference path, despite offering large, cost-effective accelerator memory pools ideal for long-sequence spatiotemporal attention. We present \texttt{vidax}, an open-source JAX/Flax inference engine and zero-copy PyTorch-to-JAX weight translator for modern video generation architectures. \texttt{vidax} covers a diverse set of spatiotemporal models --- including Diffusion Transformers, omnimodal Mixture-of-Transformers, 3D VAEs, text encoders, and native samplers --- with zero PyTorch dependency in the execution path. The framework unifies 1D tensor parallelism with DeepSpeed-Ulysses sequence parallelism on a single JAX sharding mesh, integrates TPU flash-attention kernels, and implements per-layer weight offloading to support reference resolutions that exceed single-device memory. We benchmark compile times, latency, and peak memory utilization on TPU v4-8 hardware, and document real-world numerical bugs surfaced during checkpoint translation. \texttt{vidax} is released open-source as a baseline for JAX and TPU video generation research.
\end{abstract}

%% file: sections/1_introduction.tex
\section{Introduction}
\label{sec:intro}

\textbf{Spatiotemporal constraints in video generation.}
Modern video generative models --- including Wan, Cosmos, LTX, HunyuanVideo, and CogVideoX --- represent a significant departure from 2D image synthesis in their memory and compute profiles. While underlying architectures range from standard Diffusion Transformers to omnimodal Mixture-of-Transformers, they collectively rely on spatio-temporal latent representations paired with full self-attention over multi-frame patch volumes. Patchifying temporal volumes inflates sequence lengths into tens of thousands of tokens, causing quadratic attention complexity and per-token activation footprints to dwarf raw parameter counts. Consequently, achieving native reference resolutions requires integrating sequence parallelism, optimized attention kernels, and memory-offloading strategies as fundamental execution requirements rather than optional performance enhancements.

\paragraph{The framework gap in the JAX ecosystem}
Despite the rapid evolution of video generation architectures, official reference implementations remain strictly tied to PyTorch and CUDA-centric execution ecosystems. Cloud TPU hardware offers competitive memory capacities and an expressive SPMD programming model through JAX. However, the ecosystem lacks a unified inference framework optimized for the multi-dimensional parallelisms that video sequence lengths demand. Porting these architectures to JAX introduces distinct system-level challenges: navigating memory-staging overheads during large-scale checkpoint loading, maintaining execution safety across compilation boundaries, and composing intra-layer tensor parallelism with inter-layer sequence parallelism within compiler constraints.

\textbf{Contributions.}
We present \texttt{vidax}, an open-source JAX/Flax inference engine and weight translation framework designed to bridge this gap. Our main contributions are as follows:

\begin{itemize}[leftmargin=2em,itemsep=1.5pt,topsep=0pt]
    \item \textbf{Unified Architecture Coverage:} A modular Flax library supporting diverse video generation backbones (DiTs and omnimodal models), 3D VAEs, and text conditioning towers, paired with a host-resident weight translator providing exact numerical matching without memory duplication.
    \item \textbf{Composable Multi-Axis Parallelism:} A 3-axis JAX sharding mesh unifying Megatron-style tensor parallelism with DeepSpeed-Ulysses sequence parallelism, enabling flexible trade-offs between parameter residency and activation memory.
    \item \textbf{Memory-Extended Execution Path:} Integrated Pallas flash-attention kernels and per-layer host weight offloading, allowing model inference on memory-constrained hardware.
    \item \textbf{System Evaluation \& Correctness Analysis:} A comprehensive benchmark suite evaluating compile times, latency, and HBM utilization on TPU v4-8 hardware, alongside a detailed breakdown of real-checkpoint correctness edge cases.
\end{itemize}

All code, benchmark configurations, and documentation are available open-source at \url{https://github.com/FlyingGiraffe/vidax}.

%% file: sections/2_system.tex
\section{System Architecture \& Design}
\label{sec:system}

\textbf{Host-resident weight translation.}
To handle large parameter trees, \texttt{vidax} implements zero-copy weight ingestion using JAX~\citep{jax2018} and Flax~\citep{flax2020} \texttt{nn.Module} architectures with explicit \texttt{setup()} hooks rather than \texttt{@nn.compact} declarations. This splits forward execution into independent preprocessing, block-loop, and postprocessing stages. Checkpoint weights are staged host-side as NumPy arrays rather than being instantiated directly via \texttt{jnp.array()}. Staging directly on-device defaults to an unsharded allocation on a single accelerator, causing peak memory spikes that trigger OOM errors on large DiTs. Staging on the host until a single \texttt{device\_put} call ensures parameters are allocated on-device exactly once, already correctly sharded across the target mesh.

\textbf{Spatiotemporal tensor and sequence sharding.}
\texttt{vidax.core.sharding} builds a 3-axis JAX mesh covering data, Megatron-style~\citep{megatronlm2019} 1D tensor, and DeepSpeed-Ulysses~\citep{ulysses2023} sequence parallelism. Tensor parallelism uses standard column- and row-parallel linear projections, relying on GSPMD~\citep{gspmd2021} for automated reduction insertions. However, for architectures where per-token activation states dominate memory --- such as models using per-token AdaLN modulation --- tensor parallelism alone is insufficient. Here, sequence parallelism partitions temporal tokens across transformer blocks, executing \texttt{all\_to\_all} transposes to head-sharded views exclusively for attention ops. When composing tensor and sequence parallelism inside \texttt{shard\_map}, local output shapes and \texttt{jax.lax.psum} reductions are defined manually to prevent redundant bias accumulation.

\textbf{3D RoPE, 3D VAEs, and flash attention.}
Because 3D Rotary Position Embedding (3D RoPE) conventions across time, height, and width are mathematically non-interchangeable, each model family retains its exact reference formulation. Unmasked attention ops dispatch to Pallas TPU flash-attention kernels~\citep{flashattention2022} to avoid materializing full query-key attention matrices, which scale quadratically with video sequence length. Because Pallas custom calls bypass GSPMD partitioning, multi-device flash-attention invocations are wrapped in explicit \texttt{shard\_map} calls. For models lacking native TPU kernels (\emph{e.g.}, LTX-2.5's neighborhood-attention VAE decoder), \texttt{vidax} falls back to a custom \texttt{scan}/\texttt{vmap}-based windowed-attention primitive. The 3D VAE architectures reuse the explicit \texttt{setup()} split, enabling frame-wise encoding and decoding as JIT-compiled per-chunk calls within a host Python loop.

\textbf{Per-layer weight offloading.}
For high-resolution regimes where keeping the full parameter tree resident leaves no headroom for the rest of the pipeline (\emph{e.g.}, Wan2.1 14B at 720p, whose fp32 DiT crowds out the VAE's execution buffers), \texttt{vidax} implements per-layer host-to-device offloading modeled on ZeRO-Offload~\citep{zerooffload2021}. Parameters remain host-resident, with layer blocks dynamically streamed into a fixed-shape HBM buffer via \texttt{device\_put}. A single \texttt{jax.jit}-compiled block function is reused across layers, using \texttt{donate\_argnums} to mutate buffers in place. Because host-to-device transfer latency does not fully overlap with compute, offloading is treated strictly as an optional fallback to prevent OOMs at reference resolutions rather than a default execution path.

\textbf{JIT compilation and memory ergonomics.}
To ensure predictable XLA compilation and prevent HLO tracing bloat, \texttt{vidax} enforces three system-wide rules. First, spatial and temporal sequence dimensions are kept strictly static to allow compiled execution graph reuse without re-tracing. Second, outer iteration loops --- such as multi-step sampling passes, layer offloading sweeps, and chunked VAE decoding --- remain plain host Python loops around individual JIT-compiled functions; wrapping an entire multi-step sampling pass in a single \texttt{jax.jit} traces the full loop into one HLO graph, forcing intermediate activations from all steps to coexist in memory. Third, precision downcasting (\emph{e.g.}, converting fp32 checkpoint weights to \texttt{bfloat16}) is performed on the host prior to device placement, avoiding the memory overhead of temporary dual-precision arrays on accelerator chips.

%% file: sections/3_models.tex
\section{Supported Models}
\label{sec:models}

\subsection{Architecture coverage}

In its initial release, \texttt{vidax} supports five primary model families (Table~\ref{tab:models}): Wan, Cosmos, LTX, HunyuanVideo, and CogVideoX. Most implementations are Diffusion Transformers (DiTs). However, architectures vary significantly across and within families; for example, Cosmos3 uses a dual-pathway omnimodal Mixture-of-Transformers rather than a DiT. Furthermore, several families group distinct architectures under a single brand name (\emph{e.g.}, Wan2.1 vs.\ Wan2.2, Cosmos-Predict2.5 vs.\ Cosmos3, and LTX-Video vs.\ LTX-2.5). In these cases, family boundaries reflect upstream product branding rather than architectural continuity, occasionally requiring separate codebase implementations.

\textbf{Wan} (2.1/2.2)~\citep{wan2025}: A family of 3D-RoPE DiTs conditioned on UMT5-XXL, with a CLIP-ViT-H/14 vision encoder added for Wan2.1's image-to-video model (Wan2.2 drops the CLIP branch, conditioning image-to-video by latent concatenation/substitution instead). Wan2.1 uses per-sample AdaLN modulation. Wan2.2 introduces two architectural variants: the A14B model (a two-expert MoE) and the TI2V-5B model. Both Wan2.2 variants switch to per-token AdaLN modulation, requiring the dedicated sequence-parallel strategy described in Section~\ref{sec:system}.

\textbf{Cosmos} (Predict2.5~\citep{cosmospredict2025} / Cosmos3~\citep{cosmos3_2026}): A family bridging standard DiT designs and unified omnimodal models. Cosmos-Predict2.5 is a DiT using a Qwen2.5-VL-7B text encoder with AdaLN-LoRA timestep modulation. Cosmos3 replaces this design entirely with a Mixture-of-Transformers that combines causal "understanding" and full-attention "generation" pathways into a single backbone, using additive timestep injection without AdaLN.

\textbf{LTX} (Video~\citep{ltxvideo2025} / 2.5~\citep{ltx2_2026}): A family focused on efficient video generation using custom VAE pipelines. LTX-Video is a standard T5-XXL-conditioned DiT paired with a pixel-unshuffle VAE. LTX-2.5 scales to 22B parameters, introducing a Gemma-4 12B text encoder, an ancestral (SDE) Euler sampler, and an optional neighborhood-attention VAE decoder that requires the windowed-attention kernel detailed in Section~\ref{sec:system}.

\textbf{HunyuanVideo}~\citep{hunyuanvideo2024,hunyuanvideo1_5_2025}: A family designed around multi-tower text and visual conditioning. HunyuanVideo-1.5 (8.3B) conditions on three encoders simultaneously: Qwen2.5-VL-7B for semantics, a byT5-small glyph encoder for precise text rendering, and SigLIP for image-to-video. The original 13B model is also supported: a dual-stream/single-stream MMDiT sharing HunyuanVideo-1.5's block implementation, but conditioned instead on a Llama-3-8B decoder tower plus a pooled CLIP-L vector (and the full multimodal LLaVA-Llama-3 model for image-to-video).

\textbf{CogVideoX} (1.0/1.5)~\citep{cogvideox2024}: A DiT family built on a 3D causal VAE and a joint text-video 3D full-attention mechanism for spatio-temporal modeling. Structural design varies by scale and version: the 2B model uses fixed sinusoidal positional embeddings rather than RoPE, whereas the 1.5 series introduces temporal patchification alongside a revised 3D-RoPE grid structure.

\begin{table}[t]
  \caption{Supported model families and their distinguishing details.}
  \label{tab:models}
  \centering
  \footnotesize
  \setlength{\tabcolsep}{3pt}
  \begin{tabular}{@{}p{2.2cm} p{2.0cm} p{4.0cm} p{2.8cm} p{1.9cm}@{}}
    \toprule
    Family & Sizes & Conditioning & Timestep inj. & Scheduler \\
    \midrule
    Wan2.1        & 1.3B, 14B        & UMT5-XXL, CLIP (I2V) & per-sample AdaLN     & Euler / rect.\ flow \\
    Wan2.2        & 5B, A14B (MoE)   & UMT5-XXL             & per-token AdaLN      & Euler / rect.\ flow \\
    Cosmos-Predict2.5 & 2B, 14B      & Qwen2.5-VL-7B         & per-frame AdaLN-LoRA & UniPC \\
    Cosmos3       & 16B, 4B & Qwen tokenizer, in-backbone & additive (no AdaLN) & UniPC, Karras \\
    LTX-Video 0.9.8 & 2B, 13B         & T5-XXL                & AdaLN                & Rect.\ flow \\
    LTX-2.5       & 22B & Gemma-4 12B         & cross-attn.\ AdaLN   & Ancestral Euler \\
    HunyuanVideo & 13B            & Llama-3-8B, CLIP-L, LLaVA (I2V) & AdaLN, embedded guid. & Flow matching \\
    HunyuanVideo-1.5 & 8.3B           & Qwen2.5-VL, byT5, SigLIP (I2V) & AdaLN        & Flow matching \\
    CogVideoX / 1.5 & 2B, 5B          & T5-v1.1-XXL            & AdaLN (v-pred.)      & DDIM / DPM \\
    \bottomrule
  \end{tabular}
\end{table}

\subsection{Weight import, precision, and verification}

Parameter conversion for every model is verified against official reference checkpoints using exact 1:1 key mapping and, where a reference PyTorch environment was available, direct numerical comparison of intermediate activations. Most models execute in \texttt{bfloat16}. However, we preserve target precision fixes where necessary: Wan2.1 maintains fp32 DiT weights because the reference implementation keeps its residual stream in fp32 under autocast, while LTX-2.5 retains specific fp32 AdaLN tables from its native checkpoint. Both precision choices prevent measurable quality degradation, as detailed in Appendix~\ref{sec:app:lessons}.

All five families are verified end-to-end and benchmarked (Section~\ref{sec:bench}). Verification includes bit-exact or high-correlation block-level parity checks against the original PyTorch implementation, exact 1:1 parameter-tree matches, and prompt-faithful end-to-end generation (examples shown in Figure~\ref{fig:qualitative}).

\subsection{Diffusion and Flow Matching Schedulers}

Three sampler families are implemented from scratch and shared across models: 
\begin{itemize}[leftmargin=2em,itemsep=1.5pt,topsep=0pt]
    \item \textbf{Euler Rectified-Flow Integration}: First-order flow-matching integrators~\citep{rectifiedflow2022,flowmatching2022}, including an ancestral stochastic differential equation (SDE) variant required for LTX-2.5.
    \item \textbf{Multistep UniPC}: A higher-order predictor-corrector scheduler adapted for flow matching~\citep{unipc2023}. UniPC achieves comparable sampling quality in significantly fewer steps (\emph{e.g.}, 35 steps for Cosmos versus $\sim$50 for Wan).
    \item \textbf{DDIM and DPM-Solver}: $v$-prediction discrete schedulers~\citep{ddim2020,dpmsolver2022} tailored for CogVideoX, representing the primary non-flow-matching pipeline in the library.
\end{itemize}

All three scheduler families are fully integrated into the \texttt{vidax} pipeline, allowing unified model execution across diverse mathematical formulations --- from flow-matching Euler (Wan, HunyuanVideo-1.5) and UniPC (Cosmos) to ancestral SDE Euler (LTX-2.5) and $v$-prediction DPM-Solver (CogVideoX). By standardizing scheduler interfaces in JAX, the library supports zero-overhead switching between first-order, higher-order, and stochastic samplers across all supported model architectures.

%% file: sections/4_benchmark.tex
\section{Empirical Evaluation \& Benchmarks}
\label{sec:bench}

\begin{figure}[tp]
  \centering
  \includegraphics[width=\linewidth]{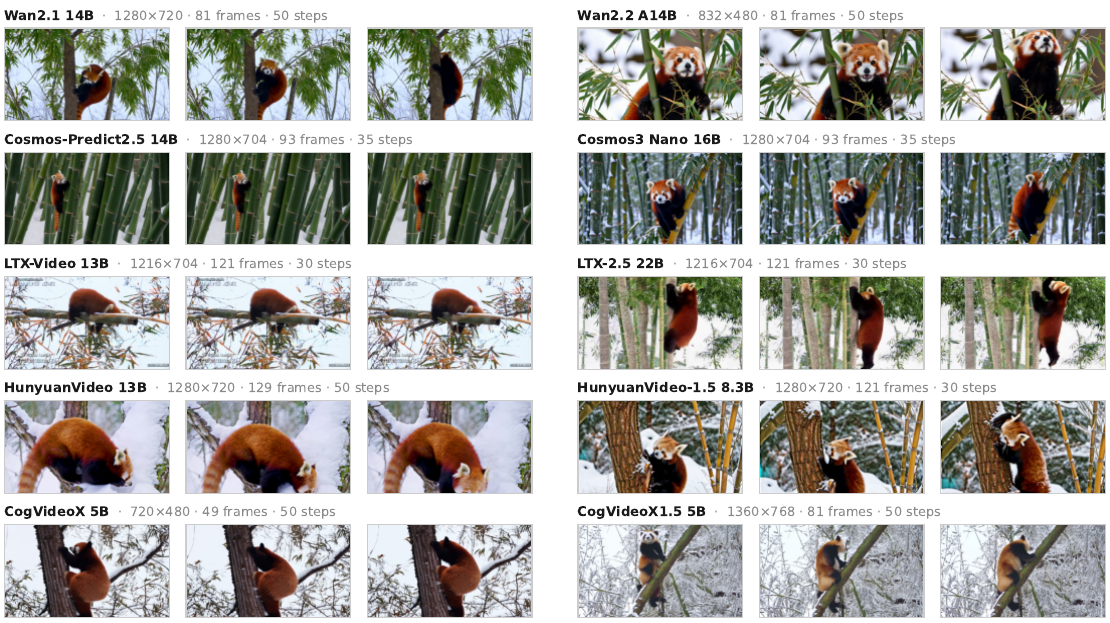}
  \caption{Generated video clips of different models; each row pairs two models from one supported family. Every clip is generated by \texttt{vidax} on TPU v4-8 from one shared standardized prompt (\emph{``A majestic red panda climbing a bamboo tree in the snow, 4k''}) at that model's reference settings.}
  \label{fig:qualitative}
\end{figure}

\textbf{Experimental setup.}
We evaluate all five model families supported by \texttt{vidax} (Table~\ref{tab:models}) across their publicly available checkpoint sizes and tasks. Benchmarks are conducted on a single TPU v4-8 slice (4 chips) running JAX v0.11.0. Each metric represents the mean of 5 independent end-to-end runs with a batch size of 1 and classifier-free guidance enabled ($2\times$ internal batch size). To measure realistic performance, resolutions, frame counts, and sampling steps use each model's reference defaults. The compilation cache is cleared prior to each run to isolate compilation overheads. Tensor parallel ($\mathrm{TP}$) and sequence parallel ($\mathrm{SP}$) configurations are noted per model.

While benchmarks focus on TPU v4-8, the codebase targets portable JAX/XLA and Pallas primitives rather than TPU-generation-specific kernels, so the same execution path carries over to newer accelerator generations like TPU v5e and v6e.

\textbf{Qualitative results.}
Figure~\ref{fig:qualitative} showcases generation results from two representative models per family, all executed via \texttt{vidax} using a standardized text prompt at default reference resolutions, frame counts, and step budgets. The evaluated pairs span key architectural variations within each family: a dense DiT versus a two-expert MoE (Wan2.1 14B vs.\ Wan2.2 A14B), a standard DiT versus a Mixture-of-Transformers (Cosmos-Predict2.5 14B vs.\ Cosmos3 Nano), and successive model generations within a family (LTX-Video 13B vs.\ LTX-2.5 22B; HunyuanVideo 13B vs.\ HunyuanVideo-1.5 8.3B; CogVideoX 5B vs.\ CogVideoX1.5 5B). Spanning five text-conditioning tower families, three scheduler formulations, and seven VAE designs, every configuration produces temporally coherent and prompt-faithful video through a single JAX execution path --- qualitatively confirming the end-to-end correctness established by our numerical and block-level verification.

\textbf{Denoising latency and memory.}
Table~\ref{tab:bench_main} summarizes compile time, per-step latency, and peak HBM utilization per chip for representative configurations at reference resolutions. A complete evaluation table of all models, tasks, and inference setups is provided in Appendix~\ref{sec:app:bench}.

\begin{table}[t]
  \caption{Representative inference performance and memory footprint on a TPU v4-8 slice (4 chips) at reference resolution, frame count, and step budget.}
  \label{tab:bench_main}
  \centering
  \footnotesize
  \setlength{\tabcolsep}{3pt}
  \begin{tabular}{lccccrrr}
    \toprule
    Model & TP/SP & Resolution & Frames & Offload & Compile (s) & Per-step (s) & Peak HBM (GB) \\
    \midrule
    Cosmos3 Nano (16B)     & 4/-- & 1280x704 & 93  & --       & 64.2  & 7.1   & 29.5 \\
    Cosmos-Predict2.5 14B  & 4/1  & 1280x704 & 93  & chunk 1  & 48.5  & 128.0 & 14.7 \\
    Wan2.2 A14B            & 2/2  & 832x480  & 81  & chunk 10 & 65.8  & 43.2  & 28.4 \\
    Wan2.1 14B             & 4/1  & 1280x720 & 81  & chunk 20 & 108.2 & 123.0 & 23.0 \\
    LTX-Video 13B (dev)    & 4/-- & 1216x704 & 121 & --       & 134.7 & 5.2   & 15.3 \\
    LTX-2.5 22B (dev)      & 4/-- & 1216x704 & 121 & chunk 8  & 87.7  & 7.3   & 16.7 \\
    HunyuanVideo-1.5 8.3B  & 4/-- & 1280x720 & 121 & --       & 410.8 & 221.0 & 30.3 \\
    HunyuanVideo 13B       & 4/-- & 1280x720 & 129 & chunk 20/40 & 506.7 & 299.6 & 18.4 \\
    CogVideoX 5B            & 4/-- & 720x480  & 49  & --       & 105.8 & 9.4   & 23.2 \\
    CogVideoX1.5 5B & 1/4 & 1360x768 & 81 & --      & 306.6 & 52.8  & 31.5 \\
    \bottomrule
  \end{tabular}
\end{table}

\begin{figure}[t]
  \centering
  \includegraphics[width=\linewidth]{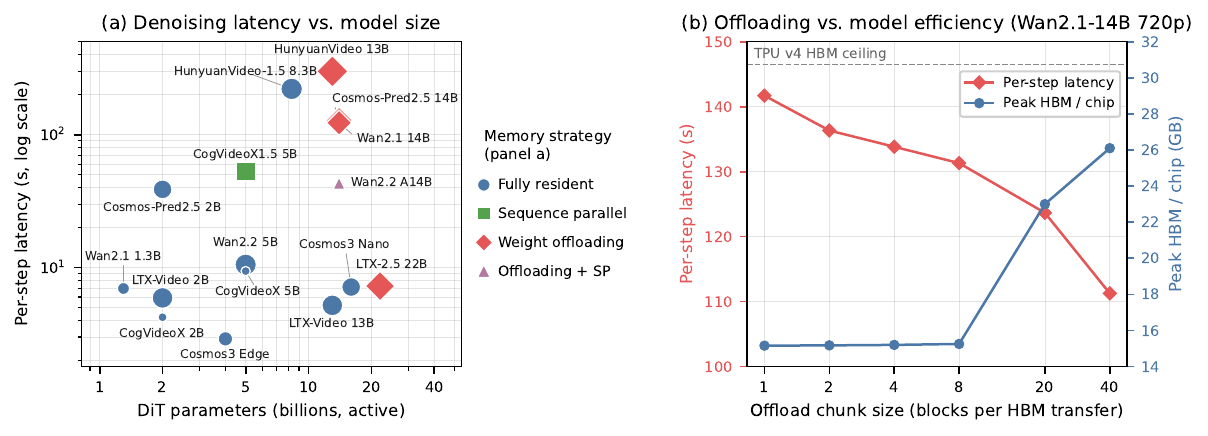}
  \caption{Performance and memory utilization on TPU v4-8. \textbf{(a)} Mean per-step latency versus DiT parameter count (active parameters), with marker area proportional to output pixel-volume and color indicating memory mitigation strategy. \textbf{(b)} Per-step latency and peak HBM per chip for Wan2.1 14B across an offload chunk-size sweep, with the dashed line denoting the usable v4 HBM ceiling.}
  \label{fig:bench_analysis}
\end{figure}

Figure~\ref{fig:bench_analysis}(a) illustrates the inference speed and memory trade-offs across model sizes. For fully device-resident configurations, per-step latency tracks parameter count and token volume predictably. Configurations requiring per-layer weight offloading (Wan2.1 14B, Cosmos-Predict2.5 14B, Wan2.2 A14B, HunyuanVideo 13B) exhibit significantly higher latency because host-to-device parameter transfers do not fully overlap with compute on this hardware. LTX-2.5 22B is an exception: its weights fit within $\mathrm{TP}=4$ HBM, and offloading is used exclusively to partition the forward trace and free intermediate activations between blocks.

\textbf{Memory optimization analysis.}
Peak HBM utilization is held close to the usable TPU v4 budget ($\sim$30.75 GB per chip) through selective tensor parallelism, sequence parallelism, and per-layer weight offloading; the most demanding configurations (CogVideoX1.5, native-720p Wan2.1 I2V) sit right at that ceiling. Without these mechanisms, native reference resolutions trigger out-of-memory errors. Table~\ref{tab:offload_summary} lists the primary memory constraints and corresponding mitigations.

\begin{table}[t]
  \caption{Primary memory constraints and required hardware mitigation strategies across models.}
  \label{tab:offload_summary}
  \centering
  \footnotesize
  \setlength{\tabcolsep}{3pt}
  \begin{tabular}{p{0.21\textwidth}p{0.10\textwidth}p{0.64\textwidth}}
    \toprule
    Model Family & Mitigation & Primary Memory Constraint \\
    \midrule
    Wan2.1 (720p) & Offload & fp32 DiT weights leave insufficient HBM for VAE execution buffers. \\
    Wan2.2 A14B & Offload+SP & High activation memory from per-token AdaLN modulation. \\
    Cosmos-Predict2.5 14B & Offload & 93-frame context does not fit fully resident at any TP/SP split. \\
    Wan2.2 5B & TP & Parameter volume fits 4-way tensor parallelism without residual pressure. \\
    LTX-Video & TP & Shards self-attention activations (13B/2B) and, for the 13B checkpoints, weights too large to fit replicated on one chip. \\
    LTX-2.5 22B & TP+Offload & TP(4) fits weights; offloading manages unfused intermediate block activations. \\
    HunyuanVideo-1.5 8.3B & TP & 4-way sharding of the DiT's own weights, needed alongside the replicated multi-tower conditioning footprint. \\
    HunyuanVideo 13B & TP+Offload & 13B weights and 129-frame 720p activations exceed HBM even at TP(4). \\
    CogVideoX1.5 5B & SP & $\sim$45k-token joint-attention activations exceed HBM under TP; requires SP=4. \\
    \bottomrule
  \end{tabular}
\end{table}

To quantify the impact of memory management, we evaluate the trade-off between resident-weight headroom and execution throughput. Per-layer weight offloading trades compute efficiency for memory capacity by streaming parameters dynamically into a fixed device buffer. As demonstrated by the chunk-size sweep on Wan2.1 14B at native 720p (Figure~\ref{fig:bench_analysis}(b)), increasing the chunk size improves transfer-compute overlap and reduces per-step latency --- falling monotonically from $141.7$ s at a chunk size of 1 to $111.3$ s when the full 40-block model is resident --- while scaling peak HBM usage from $15.2$ GB to $26.1$ GB. Consequently, weight offloading serves strictly as a memory-fit mechanism for high-resolution execution paths rather than a throughput-neutral optimization.

%% file: sections/5_conclusions.tex
\section{Conclusions \& Future Work}
\label{sec:conclusion}

We presented \texttt{vidax}, a JAX/Flax inference engine and weight translator that brings major open-source video generation models to Cloud TPUs with zero PyTorch dependency during execution. By combining tensor and sequence parallelism, TPU flash-attention kernels, and per-layer weight offloading, the engine successfully runs reference-resolution video generation within device memory constraints. Additionally, the porting process identified concrete numerical and structural edge cases across official checkpoints that pure simulation studies often miss.

Future development may focus on three primary directions:
\begin{itemize}[leftmargin=2em,itemsep=1.5pt,topsep=0pt]
\item \textbf{Hardware generalization:} Extending benchmarking and performance validation beyond TPU v4 slices to newer TPU architectures (v5e and v6e).
\item \textbf{Training and fine-Tuning:} Leveraging the existing sharded execution graph and offloading infrastructure to support post-training workloads, such as LoRA and full-parameter fine-tuning.
\item \textbf{Low-level kernel optimization:} Developing custom Pallas and Mosaic kernels tailored to the TPU memory hierarchy to reduce latency overheads.
\end{itemize}

%% file: sections/X_appendix.tex
\begin{table}[htbp]
  \caption{Taxonomy of correctness and performance bugs identified during cross-framework porting.}
  \label{tab:lessons}
  \centering
  \footnotesize
  \setlength{\tabcolsep}{4pt}
  \begin{tabular}{p{0.20\textwidth}p{0.37\textwidth}p{0.37\textwidth}}
    \toprule
    \textbf{Symptom} & \textbf{Root Cause} & \textbf{Resolution} \\
    \midrule
    \multicolumn{3}{l}{\textit{\textbf{1. Precision, Rounding \& Accumulation Errors}}} \\
    \midrule
    Wan2.1 14B I2V: hazy, low-detail output at native 720p/81f.
      & Running residual additions in \texttt{bf16} compounds rounding errors across the 40-block chain at high token counts.
      & Use \texttt{fp32} for DiT parameters/accumulators; cast to \texttt{bf16} transiently only during matmuls. \\
    \addlinespace
    CogVideoX: 16--37\% T5 embedding drift vs. PyTorch reference.
      & T5-XXL residual stream reaches $\sim 10^5$; attending unmasked padding tokens compounds \texttt{bf16} rounding across 24 layers.
      & Execute T5 forward pass in \texttt{fp32}, cast outputs to \texttt{bf16}, and immediately free \texttt{fp32} weights. \\
    \addlinespace
    LTX-Video: port outputs agree with PyTorch only to $\sim$2 decimals.
      & JAX defaults to lower-precision TPU matrix multiplication (\texttt{"default"}) for throughput.
      & Set \texttt{jax\_default\_matmul\_precision="highest"} during parity checks (reduces max diff to $\sim 3\times 10^{-5}$). \\
    \addlinespace
    LTX-2.5 (dev): washed-out output, blown highlights, and compounding quality loss.
      & Missing guidance-rescale correction caused CFG over-saturation; a blanket dtype cast also downcast the 290 AdaLN modulation tables the checkpoint ships in \texttt{fp32}.
      & Add \texttt{--guidance\_rescale} (0.7 for dev); preserve \texttt{fp32} for \texttt{scale\_shift\_table} parameter leaves regardless of \texttt{--dit\_dtype}. \\

    \midrule
    \multicolumn{3}{l}{\textit{\textbf{2. Sharding, Parallelism \& Custom Kernels}}} \\
    \midrule
    Wan2.1/2.2: blocky output under Weight-Offload + SP.
      & Row-parallel \texttt{psum} sums fully replicated \texttt{nn.Dense} biases once per TP device.
      & Implement \texttt{psum\_row\_parallel}: subtract bias, perform \texttt{psum}, and add a single bias copy back. \\
    \addlinespace
    HunyuanVideo-1.5: \texttt{RESOURCE\_EXHAUSTED} at 720p resolution.
      & Dense attention mask materialization $(B, H, S_q, S_k)$ bypasses Pallas flash attention $O(S)$ memory savings.
      & Replaced dense mask tensor with Pallas \texttt{SegmentIds} to skip invalid key blocks dynamically. \\
    \addlinespace
    HunyuanVideo-1.5: Mosaic partitioning failure with replicated ops.
      & Mixing sharded parameters with unsharded kernels (\texttt{SingleTokenRefiner}) breaks global JIT trace.
      & Wrap unsharded kernel dispatches in an explicit fully replicated \texttt{shard\_map}. \\

    \midrule
    \multicolumn{3}{l}{\textit{\textbf{3. Model Architecture \& Conditioning Parity}}} \\
    \midrule
    Cosmos-Predict2.5: color grid artifacts across all settings.
      & Incorrect EDM preconditioning wrapper, unpatchify channel swap, and missing timestep rescaling.
      & Removed preconditioning, corrected channel ordering, and restored internal DiT timestep scaling. \\
    \addlinespace
    Cosmos3 Edge (4B): flat, featureless output on short prompts.
      & Edge (4B) strictly requires JSON-structured expanded prompts, whereas Nano (16B) tolerates raw text.
      & Enforce checkpoint-native JSON prompt formatting during tokenization. \\
    \addlinespace
    LTX-2.5: semantic prompt drift across random seeds.
      & Feature-extractor scaling used total 49-layer width as denominator instead of Gemma's single-layer width.
      & Scale conditioning vectors by single-layer \texttt{embedding\_dim}. \\
    \addlinespace
    HunyuanVideo: Llama encoder divergence on padded prompts.
      & Tokenizer default \texttt{padding\_side} diverged from reference, invalidating left/right causal assumptions.
      & Pass \texttt{padding\_side="right"} explicitly to \texttt{AutoTokenizer}. \\

    \midrule
    \multicolumn{3}{l}{\textit{\textbf{4. Memory, Tiling \& Execution Graphs}}} \\
    \midrule
    LTX-2.5 VAE Decoder: 147\,GB HLO temporaries at a tiny synthetic latent.
      & Naive local-window attention materializes memory proportional to the product of all window axes ($11^3$ for this checkpoint's stage-5 kernel).
      & Rebuilt 3D neighborhood attention around \texttt{lax.scan} (one query row at a time) plus \texttt{vmap} for the inner gather. \\
    \addlinespace
    LTX-2.5 VAE Decoder: compile appears to hang ($>$1 hour) at real resolution.
      & A first attempt to fix the OOM above (looping per query step in Python) unrolls at trace time into near-identical subgraph copies across every block.
      & \texttt{lax.scan} compiles the per-step body once regardless of step count; final $\mathrm{tp}=4$ sharding then resolves a remaining OOM down to 14.76\,GB/chip. \\
    \addlinespace
    CogVideoX: boundary artifacts at VAE spatial tile seams.
      & Initial implementation blended each tile against the pristine, unmodified neighbor grid, unlike the reference's stateful blend.
      & Match the reference: write each blended tile back into the grid in place before it is used as a neighbor for the next. \\
    \bottomrule
  \end{tabular}
\end{table}

\section{Debugging Notes and System Verification}
\label{sec:app:lessons}

Validating \texttt{vidax} against real, pre-trained video generation checkpoints exposed several correctness edge cases and performance bottlenecks that remain invisible during synthetic testing or unsharded execution. Table~\ref{tab:lessons} categorizes these issues by their root systems causes across numerical precision, parallel execution, architecture parity, and memory graph boundaries.

\textbf{Diagnostic Methodologies.}
Two specific debugging practices proved essential for isolating errors during engine development:
\begin{itemize}[leftmargin=1.5em,itemsep=2pt,topsep=2pt]
    \item \textbf{Low-Noise Real-Photo Probes:} When diffusion models emit unstructured noise or grid artifacts, standard forward passes cannot easily distinguish between incorrect noise-level conditioning and corrupted network weights. By encoding a real image into latent space, adding a small amount of noise, executing a single denoising step, and decoding, we isolated a noise-level-conditioning (preconditioning) bug from a general weights/attention fault within a single iteration.
    \item \textbf{Explicit Precision Auditing:} Mixed-precision checkpoints must be audited at the individual tensor level rather than relying on global type casting. Preserving \texttt{fp32} precision on specific parameters (such as AdaLN tables in LTX-2.5 and residual accumulators in Wan2.1) was crucial for numerical parity.
\end{itemize}

\textbf{Systems Insights on Declarative Sharding.}
A key architectural takeaway from this implementation concerns JAX's sharding abstraction model. Using \texttt{jax.sharding} declarations alongside GSPMD auto-partitioning simplifies Megatron-style tensor parallelism down to annotating parameter trees, eliminating manual collective operations (\emph{e.g.}, \texttt{all-reduce}). However, this automation ends at the boundary of custom kernels (\texttt{shard\_map} and Pallas dispatch). At this boundary, manual communication logic must be explicitly managed, highlighting a fundamental abstraction line in current JAX systems engineering.

\section{Full benchmark table}
\label{sec:app:bench}

Table~\ref{tab:bench_full} provides the complete benchmark evaluation across all supported DiT inference configurations measured on a single TPU v4-8 slice (4 chips, \texttt{jax==0.11.0}). The remainder of this section details the experimental methodologies, architectural trade-offs, and memory mitigation mechanisms underlying these measurements.

\begin{table}[htbp]
  \caption{Performance and memory footprint across all evaluated model configurations on
TPU v4-8.}
  \label{tab:bench_full}
  \centering
  \footnotesize
  \setlength{\tabcolsep}{2pt}
  \resizebox{\textwidth}{!}{%
  \begin{tabular}{llcccccclrrr}
    \toprule
    Model & Variant & Task & TP/SP & Res. & Frames & Steps & Weight & Offload & Compile (s) & Per-step (s) & HBM (GB) \\
    \midrule
    Cosmos3 & Nano (16B) & T2V & 4/-- & 1280x704 & 93 & 35 & bf16 & -- & 64.2 & 7.1 & 29.5 \\
    Cosmos3 & Edge (4B) & T2V & 4/-- & 832x480 & 121 & 35 & bf16 & -- & 64.5 & 2.4 & 17.0 \\
    Cosmos-Predict2.5 & 14B & T2V & 4/1 & 1280x704 & 93 & 35 & bf16 & chunk 1 & 48.5 & 128.0 & 14.7 \\
    Cosmos-Predict2.5 & 2B & T2V & 4/1 & 1280x704 & 93 & 35 & bf16 & -- & 112.9 & 38.8 & 16.0 \\
    Wan2.2 & A14B & T2V & 2/2 & 832x480 & 81 & 50 & fp32 & chunk 10 & 65.8 & 43.2 & 28.4 \\
    Wan2.2 & A14B & T2V & 2/2 & 1280x720 & 33 & 50 & fp32 & chunk 1 & 33.7 & 46.4 & 18.1 \\
    Wan2.2 & A14B & I2V & 2/2 & 544x720$^*$ & 81 & 40 & fp32 & chunk 10 & 146.1 & 44.5 & 28.3 \\
    Wan2.2 & A14B & I2V & 2/2 & 832x1104$^*$ & 33 & 40 & fp32 & chunk 1 & 102.9 & 49.1 & 20.5 \\
    Wan2.2 & 5B & T2V & 4/1 & 1280x704 & 121 & 50 & fp32 & -- & 87.3 & 10.5 & 18.3 \\
    Wan2.2 & 5B & I2V & 4/1 & 704x1280$^*$ & 121 & 40 & fp32 & -- & 145.8 & 12.1 & 18.3 \\
    Wan2.1 & 14B & T2V & 4/1 & 1280x720 & 81 & 50 & fp32 & chunk 20 & 108.2 & 123.0 & 23.0 \\
    Wan2.1 & 14B & T2V & 4/1 & 832x480 & 81 & 50 & bf16 & -- & 142.5 & 26.1 & 17.2 \\
    Wan2.1 & 14B (720P) & I2V & 4/1 & 832x1104$^*$ & 81 & 40 & fp32 & chunk 20 & 131.3 & 127.2 & 32.7 \\
    Wan2.1 & 14B (480P) & I2V & 4/1 & 544x720$^*$ & 81 & 40 & bf16 & -- & 150.3 & 28.1 & 22.1 \\
    Wan2.1 & 1.3B & T2V & 4/1 & 832x480 & 81 & 50 & bf16 & -- & 85.4 & 7.0 & 10.2 \\
    LTX-2.5 & 22B (dev) & T2V & 4/-- & 1216x704 & 121 & 30 & bf16$^\dagger$ & chunk 8 & 87.7 & 7.3 & 16.7 \\
    LTX-2.5 & 22B (distilled) & T2V & 4/-- & 1216x704 & 121 & 8 & bf16$^\dagger$ & chunk 8 & 87.9 & 4.7 & 15.3 \\
    LTX-2.5 & 22B (dev), diff.\ VAE & T2V & 4/-- & 1216x704 & 121 & 30 & bf16$^\dagger$ & chunk 8 & 479.5 & 95.3$^\ddagger$ & 16.1 \\
    LTX-2.5 & 22B (distilled), diff.\ VAE & T2V & 4/-- & 1216x704 & 121 & 8 & bf16$^\dagger$ & chunk 8 & 478.2 & 335.1$^\ddagger$ & 14.8 \\
    LTX-Video (0.9.8) & 13B (dev) & T2V & 4/-- & 1216x704 & 121 & 30 & bf16 & -- & 134.7 & 5.2 & 15.3 \\
    LTX-Video (0.9.8) & 13B (distilled) & T2V & 4/-- & 1216x704 & 121 & 8 & bf16 & -- & 136.4 & 13.0 & 15.3 \\
    LTX-Video (0.9.8) & 2B (distilled) & T2V & 4/-- & 1216x704 & 121 & 8 & bf16 & -- & 83.5 & 5.9 & 8.8 \\
    HunyuanVideo-1.5 & 8.3B (720P) & T2V & 4/-- & 1280x720 & 121 & 30 & bf16 & -- & 410.8 & 221.0 & 30.3 \\
    HunyuanVideo-1.5 & 8.3B (720P) & I2V & 4/-- & 832x1104$^*$ & 121 & 30 & bf16 & -- & 416.7 & 219.2 & 32.0 \\
    HunyuanVideo-1.5 & 8.3B (480P) & T2V & 4/-- & 832x480 & 121 & 30 & bf16 & -- & 362.3 & 119.5 & 29.1 \\
    HunyuanVideo-1.5 & 8.3B (480P) & I2V & 4/-- & 544x720$^*$ & 121 & 30 & bf16 & -- & 362.4 & 112.9 & 31.4 \\
    HunyuanVideo & 13B & T2V & 4/-- & 1280x720 & 129 & 50 & bf16 & chunk 20/40 & 506.7 & 299.6 & 18.4 \\
    HunyuanVideo & 13B & I2V & 4/-- & 832x1088$^*$ & 129 & 50 & bf16 & chunk 20/40 & 557.3 & 306.4 & 18.4 \\
    CogVideoX1.5 & 5B & T2V & 1/4 & 1360x768$^\parallel$ & 81 & 50 & bf16 & -- & 306.6 & 52.8 & 31.5 \\
    CogVideoX1.5 & 5B & I2V & 1/4 & 1360x768$^{\parallel\P}$ & 81 & 50 & bf16 & -- & 304.2 & 52.8 & 31.5 \\
    CogVideoX & 5B & T2V & 4/-- & 720x480 & 49 & 50 & bf16 & -- & 105.8 & 9.4 & 23.2 \\
    CogVideoX & 5B & I2V & 4/-- & 720x480$^\P$ & 49 & 50 & bf16 & -- & 106.1 & 9.4 & 23.3 \\
    CogVideoX & 2B & T2V & 2/-- & 720x480 & 49 & 50 & bf16$^\S$ & -- & 48.4 & 4.2 & 17.2 \\
    \bottomrule
  \end{tabular}%
  }
\end{table}

\textbf{Experimental Methodology.}
Each reported row represents the mean across 5 independent end-to-end runs with a batch size of 1 and classifier-free guidance enabled ($2\times$ internal batch size). Prior to each run, the XLA compilation cache is cleared to isolate compilation latency from execution runtime. All runs utilize a default \texttt{bf16} input/output dtype across activations, VAE, and text encoders. The \textit{Weight} column denotes the model-specific storage dtype for the DiT parameters. Resolutions, frame counts, and step counts reflect each model family's default reference specifications. Benchmark evaluations use standardized text prompts (and conditioning images for I2V tasks) to maintain cross-family comparability.

\textbf{Metrics and Latency Decomposition.}
Performance metrics are split into one-time compilation overheads and sampling runtime:
\begin{itemize}[leftmargin=1.5em,itemsep=2pt,topsep=2pt]
    \item \textbf{Compile (s):} Measures the single execution overhead required by XLA to trace and compile the model graph for a fixed tensor shape, precision, and sharding specification. This cost is incurred once per execution signature and amortized over long-running serving deployments.
    \item \textbf{Per-step (s):} Calculated as total generation latency (sampling loop plus VAE decoding) divided by the step budget. This serves as the primary standardized metric for evaluating execution speed across models with differing sampling steps. For LTX-2.5's diffusion-VAE-decoder rows ($^\ddagger$), this figure is dominated by a one-time VAE-decode compile and execution rather than the DiT sampling cost, so it is not directly comparable across the two VAE variants.
    \item \textbf{Peak HBM (GB):} Represents the maximum memory allocation watermark per chip. Keeping this value at or near the usable TPU v4 limit ($\sim$30.75\,GB per chip) dictates the necessary sharding and offloading strategy; the most demanding rows (\emph{e.g.}, native-720p Wan2.1 I2V, CogVideoX1.5) sit slightly above it.
\end{itemize}

\textbf{Sharding and Weight Offloading Dynamics.}
Configurations specify parameter parallelism via $\mathrm{TP}$ (tensor parallelism) and $\mathrm{SP}$ (sequence parallelism). Where memory limits prevent device-resident execution, per-layer weight offloading streams parameter blocks dynamically into a static device buffer:
\begin{itemize}[leftmargin=1.5em,itemsep=2pt,topsep=2pt]
    \item \textbf{Offloading Offsets for Auxiliary Stages:} Wan2.1 (720p) uses offloading to preserve HBM headroom for the VAE's execution buffers, and Cosmos-Predict2.5 (14B) uses it because its full 93-frame context does not fit fully resident at any $\mathrm{TP}/\mathrm{SP}$ split --- in both cases even though the raw DiT parameters fit on-chip. Adjusting the offloading chunk size allows tuning the trade-off between transfer-compute overlap and peak memory utilization.
    \item \textbf{Combined Offloading and Sequence Parallelism:} Wan2.2 A14B uses per-token AdaLN modulations, making activation memory the binding bottleneck rather than parameter volume. Consequently, it shards the currently-resident expert with both tensor parallelism ($\mathrm{TP}=2$) and sequence parallelism ($\mathrm{SP}=2$), plus per-layer offloading for the remaining HBM headroom. Conversely, Wan2.2 5B manages memory pressure using pure tensor parallelism ($\mathrm{TP}=4$).
    \item \textbf{Activation Memory Truncation in LTX Models:} LTX-Video relies on 4-way tensor parallelism ($\mathrm{TP}=4$) primarily to shard self-attention activations over 121 frames. LTX-2.5 22B combines $\mathrm{TP}=4$ with weight offloading (\texttt{chunk 8}); here, offloading functions as a graph-splitting mechanism to free intermediate activations between transformer blocks rather than purely as a parameter storage solution. Its Weight column reports the dominant \texttt{bf16} dtype ($^\dagger$); a small set of AdaLN modulation tables is preserved at \texttt{fp32} regardless of \texttt{-{}-dit\_dtype}.
\end{itemize}

\textbf{Precision and Architecture-Specific Execution Edge Cases.}
\begin{itemize}[leftmargin=1.5em,itemsep=2pt,topsep=2pt]
    \item \textbf{Wan2.1 Precision Requirements:} Wan2.1's native-720p rows enforce \texttt{fp32} DiT parameter precision (\texttt{--dit\_dtype float32}, the reference-matching default) to match the reference implementation's residual accumulation behavior, as \texttt{bf16} parameter quantization introduces noticeable output degradation over long token sequences at that scale; the smaller 480p/1.3B rows shown here use \texttt{bf16} safely instead.
    \item \textbf{HunyuanVideo Full Attention Overhead:} HunyuanVideo-1.5 shards parameters across $\mathrm{TP}=4$ alongside a replicated Qwen2.5-VL model. Due to unwindowed global attention across joint image-text sequences, its 720p execution exhibits higher step latency (221.0\,s/step). HunyuanVideo 13B specifies dual offload chunks (\texttt{chunk 20/40}) to independently stream its 20 double-stream and 40 single-stream blocks.
    \item \textbf{CogVideoX Parallelism Constraints:} CogVideoX1.5 5B executes at $1360{\times}768$ ($^\parallel$) using DeepSpeed-Ulysses sequence parallelism ($\mathrm{SP}=4$), as plain tensor parallelism fails compilation on a TPU v4-8 slice. CogVideoX 2B employs $\mathrm{tp}=2$ because its 30 attention heads are indivisible by 4, and its checkpoint ships as fp16 ($^\S$), cast to \texttt{bf16} here for column comparability.
    \item \textbf{I2V Dynamic Resolutions:} Image-to-Video ($^*$) models adjust target spatial dimensions dynamically based on input aspect ratios and maximum token area constraints rather than fixed square/landscape dimensions; this does not apply to the CogVideoX I2V rows below ($^\P$).
    \item \textbf{Fixed-Resolution I2V Exceptions ($^\P$):} CogVideoX 5B I2V and CogVideoX1.5 5B I2V are each locked by a learned positional-embedding buffer to their T2V sibling's fixed resolution ($720{\times}480$ / $1360{\times}768$) rather than deriving it from the conditioning image; the image is resized into that fixed box, generation proceeds there, and by default the output is rescaled back to the conditioning image's own aspect ratio afterward.
\end{itemize}